%% file: main/iclr2025_conference.tex
\documentclass{article} 
\usepackage{iclr2025_conference,times}

\input{math_commands.tex}

\usepackage{bm}
\usepackage{amssymb}
\usepackage{amsmath}
\usepackage{amsthm}
\usepackage{bbding}
\usepackage{pifont}
\usepackage{multirow}
\usepackage{bbm}

\usepackage{wrapfig}
\usepackage{subfig}
\usepackage[super]{nth}
\usepackage{float}
\usepackage{algorithm}
\usepackage{algorithmicx}
\usepackage[noend]{algpseudocode}
\usepackage{hyperref}       
\usepackage[capitalize]{cleveref}
\usepackage{color, colortbl}
\usepackage{authblk}

\usepackage[utf8]{inputenc} 
\usepackage[T1]{fontenc}    
\usepackage{graphicx}
\usepackage{url}            
\usepackage{booktabs}       
\usepackage{amsfonts}       
\usepackage{nicefrac}       
\usepackage{microtype}      
\usepackage{xcolor}         

\title{InfiniteTalk: Audio-driven Video Generation for Sparse-Frame Video Dubbing}

\newcommand*\samethanks[1][\value{footnote}]{\footnotemark[#1]}
\author[1, 2, 3]{Shaoshu Yang\thanks{Equal contribution}\;\;}
\author[4, 2, 5]{Zhe Kong\samethanks\;\;}
\author[2]{Feng Gao\samethanks\;\;}
\author[2]{Meng Cheng\samethanks\;\;}
\author[6, 2, 1]{Xiangyu Liu\samethanks\;\;}
\author[2]{\\Yong Zhang\thanks{Corresponding author}\;\;}
\author[2]{Zhuoliang Kang}
\author[5]{Wenhan Luo}
\author[2]{Xunliang Cai}
\author[1, 3]{Ran He}
\author[2]{Xiaoming Wei}

\affil[1]{School of Artificial Intelligence, University of Chinese Academy of Sciences}
\affil[2]{Meituan}
\affil[3]{New Laboratory of Pattern Recognition (NLPR), CASIA}
\affil[4]{Shenzhen Campus of Sun Yat-sen University}
\affil[5]{Division of AMC and Department of ECE, HKUST}
\affil[6]{State Key Laboratory of Multimodal Artificial Intelligence Systems, CASIA}

\affil[]{\url{https://github.com/MeiGen-AI/InfiniteTalk}}

\iclrfinalcopy 

\begin{document} 

\maketitle
\begin{center}
\captionsetup{type=figure}
  \includegraphics[width=1.0\textwidth]{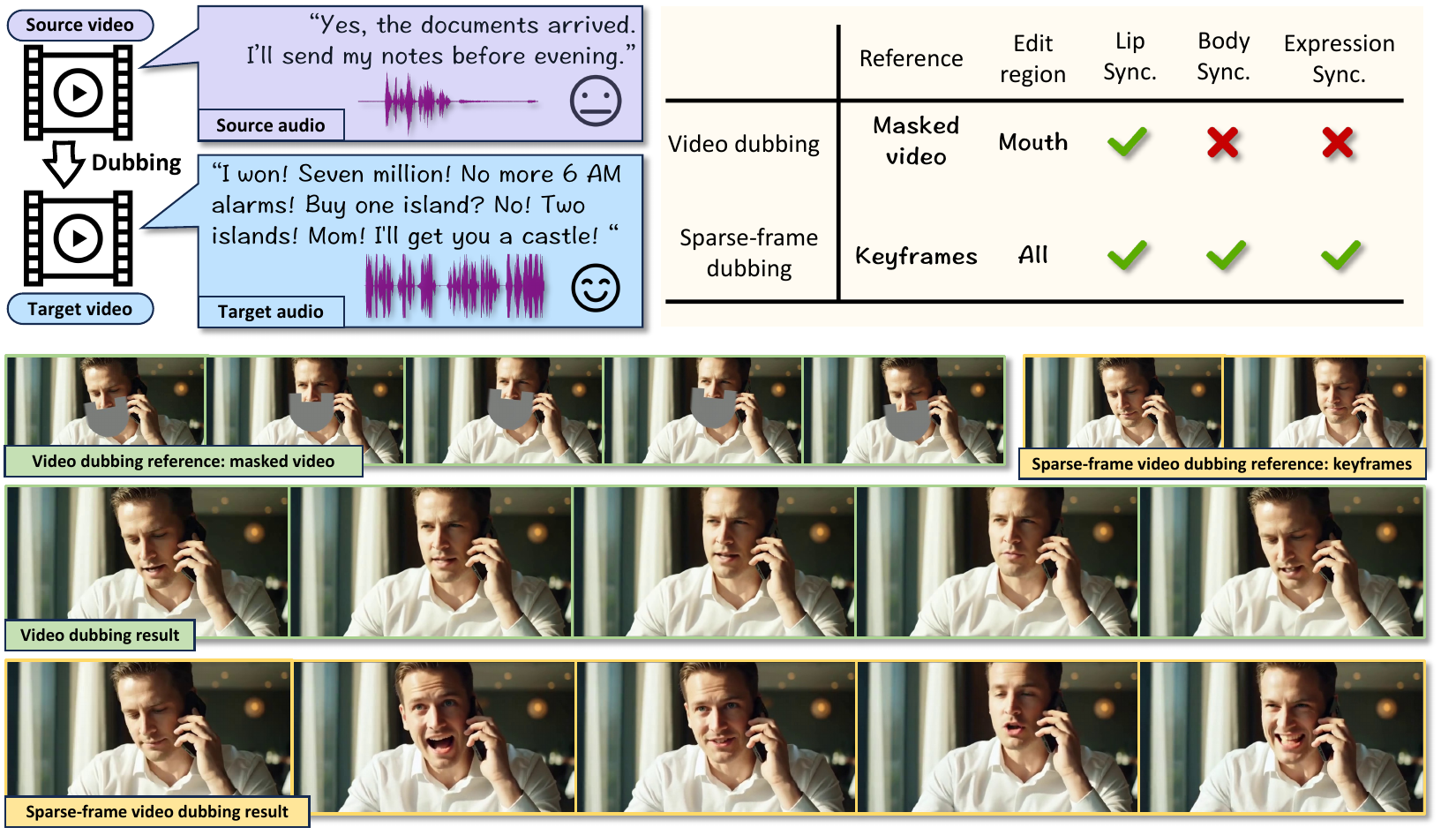}
  \captionof{figure}{
     Compared to the traditional paradigm, sparse-frame video dubbing will not only edit mouth regions. It gives the model freedom to generate audio aligned mouth, facial, and body movements while referencing on sparse keyframes to preserve identity, emotional cadence, and iconic gestures. 
  }
  \label{fig:teaser}
\end{center}

\input{sec/0_abstraction}
\input{sec/1_introduction}

\input{sec/2_related_works}
\input{sec/3_method}
\input{sec/4_experiment}
\input{sec/5_conclusion}



\bibliography{iclr2025_conference}
\bibliographystyle{iclr2025_conference}

\end{document}

%% file: math_commands.tex
\usepackage{amsmath,amsfonts,bm}

\def\eqref#1{equation~\ref{#1}}

\def\1{\bm{1}}

\DeclareMathAlphabet{\mathsfit}{\encodingdefault}{\sfdefault}{m}{sl}
\SetMathAlphabet{\mathsfit}{bold}{\encodingdefault}{\sfdefault}{bx}{n}



%% file: sec/0_abstraction.tex
\begin{abstract}
Recent breakthroughs in video AIGC have ushered in a transformative era for audio-driven human animation. However, conventional video dubbing techniques remain constrained to mouth region editing, resulting in discordant facial expressions and body gestures that compromise viewer immersion. To overcome this limitation, we introduce sparse-frame video dubbing—a novel paradigm that strategically preserves reference keyframes to maintain identity, iconic gestures, and camera trajectories while enabling holistic, audio-synchronized full-body motion editing. Through critical analysis, we identify why naive image-to-video models fail in this task, particularly their inability to achieve adaptive conditioning. Addressing this, we propose InfiniteTalk: a streaming audio-driven generator designed for infinite-length long sequence dubbing. This architecture leverages temporal context frames for seamless inter-chunk transitions and incorporates a simple yet effective sampling strategy that optimizes control strength via fine-grained reference frame positioning. Comprehensive evaluations on HDTF, CelebV-HQ, and EMTD datasets demonstrate state-of-the-art performance. Quantitative metrics confirm superior visual realism, emotional coherence, and full-body motion synchronization.
\end{abstract}

%% file: sec/1_introduction.tex
\section{Introduction}

Video dubbing is an audio-driven video-to-video generation task that combines an original video with new audio to create localized content \cite{li2024latentsync, musetalk, keysync}. This process requires editing facial movements, head rotations, and body gestures to synchronize with the dubbed speech's timing and emotional tone, while preserving the source video's visual style and camera motion. These capabilities are essential for global media distribution through streaming platforms.

Recent advances in audio-driven generative models have significantly improved lip synchronization capabilities for video dubbing \cite{li2024latentsync}. However, these methods predominantly focus on oral region inpainting, resulting in mismatched head rotations and body gestures that undermine viewer immersion. To address this limitation, we introduce sparse-frame video dubbing, a novel paradigm that preserves only reference keyframes while leveraging modern generative foundation models. As shown in \cref{fig:teaser}, it references only select keyframes to preserve the original video's emotional cadence, symbolic gestures, and camera trajectories, while liberating facial expressions, head motions, and body dynamics to synchronize organically with dubbed audio. As a long video generation task, it demands robust temporal continuation capabilities. A practical solution is employing audio-conditioned video generators with initial and terminal frame guidance.

Unfortunately, naive application of audio-conditioned image-to-video generators produces unsatisfactory dubbing results, as shown in \cref{fig:observation1}. These models fundamentally struggle with identity preservation during extended generation and default to strict motion copying on the conditioning frames. This results in stiff facial expressions, lip and head movements that contradicts speech dynamics. Meanwhile, simply applying initial and terminal frame conditioning creates abrupt inter-chunk transitions. 





To resolve these challenges, we introduce InfiniteTalk, a audio-driven generator for long sequence sparse-frame video dubbing. It has a streaming video generation base that utilize context frames to inject momentum information that creates smooth inter-chunk transitions. To preserve human identity, background, and camera movements of the source video, it controls the output video by referencing keyframes. To achieve the soft reference mechanism in sparse-frame video dubbing, we investigate how and find the control strength is determined by the similarity between the video context and image condition. Based on our investigation, we propose a sampling strategy that balances control strength and motion alignment by fine-grained reference frame positioning, achieving high quality infinite length long sequence video dubbing with full body audio-aligned motion generation. Meanwhile, we explore methods to achieve accurate subtle camera movement preservation.

To comprehensively evaluate our method, we conduct quantitative and qualitative experiments on HDTF \cite{hdtf}, CelebV-HQ \cite{celebvhq}, and EMTD \cite{emtd}, including the cases for both face and full body animation. The quantitative results show InfiniteTalk achieves the state-of-the-art performance in audio synchronized motion generation and visual quality. Our human evaluation shows InfiniteTalk successfully produces plausible lip, face, and body movements that align with the speech cadence and the emotional expression. We finish an ablation experiment concerning the sampling strategy and the strength of control, showcasing the effectiveness of our algorithm design.


We summarize our contributions as follows: (1) We introduce sparse-frame video dubbing, a novel paradigm for human-centric audio-driven video-to-video generation to produce natural facial expressions, head motions, and body dynamics that synchronize organically with dubbed audio. (2) We analyze the reasons why audio-driven image-to-video generators fail to achieve satisfying performance in this task and how the reference frame positioning during training determines the control strength. With these observations, we propose propose InfiniteTalk, a streaming long video generator with soft conditioning training strategy. (3) Extensive experiments show our method achieves the state-of-the-art performance for video dubbing, especially in lip, head, body motion synchronization.

%% file: sec/2_related_works.tex
\section{Related works}
\subsection{Video Generation}

Recent advances in generative learning methods—including autoregressive models \cite{var}, diffusion models \cite{ddnm, ddim, im-ddpm}, and flow matching \cite{flowmatching}—have revolutionized video generation. Early efforts like the Video Diffusion Model \cite{video-dm} pioneered pixel-space denoising, while later works (Make-A-Video \cite{make-a-video}, PYoCo \cite{pyoco}, Imagen Video \cite{imagen-video}) integrated large language models for text-to-video synthesis. To tackle video dimensionality, research shifted toward latent space learning: VideoGPT \cite{VideoGPT} combined VQ-VAE \cite{vq-gan} with transformers, establishing foundational latent modeling. Subsequent innovations employed diffusion models to approximate latent distributions, spawning latent video diffusion frameworks (\cite{lvdm, magicvideo, make-your-vid, video-ldm, stable-video-diffusion, wang2023lavie, chen2023videocrafter1, chen2024videocrafter2}). Among these, CogVideoX \cite{yang2024cogvideox} introduced diffusion transformers and temporal-compressed VAEs, enhancing motion complexity. Leveraging expanded datasets and compute, modern large-scale generators (\cite{genmo2024mochi, kong2024hunyuanvideo, wan}) now achieve unprecedented quality.

\subsection{Audio-driven Human Animation}

Audio-driven human animation aims to generate dynamic videos from a static reference image, producing synchronized facial expressions and body movements based on audio control signals. In recent years, with the success of diffusion models, end-to-end audio-to-video synthesis methods \cite{tian2024emo,wei2024aniportrait,xu2024hallo,chen2025echomimic,cui2024hallo3,ji2024sonic,li2024latentsync,jiang2024loopy} have demonstrated considerable potential. These approaches eliminate the need for intermediate representations and exhibit superior performance in portrait animation for talking head generation. Another line of research extends talking head generation to talking body generation. Methods in this category \cite{lincyberhost,tian2025emo2,lin2025omnihuman,meng2024echomimicv2,gan2025omniavatar,wang2025fantasytalking} have achieved significant progress, demonstrating enhanced naturalness and consistent portrait animation capabilities, largely attributed to large-scale, high-quality training data. Recently, some works \cite{chen2025hunyuanvideo,kong2025let,huang2025bind} have attempted to advance from single human animation to multi-human animation.

The aforementioned human animation approaches have achieved satisfactory results in synthesizing short videos. However, when generating longer video sequences, they encounter error accumulation issues \cite{kong2025dam}, such as identity degradation and color deviations. A concurrent work, named StableAvatar \cite{stableavatar}, also achieves infinite-long sequence human animation. But it uses only a single image as the condition, and is not capable for the video-to-video task.

%% file: sec/3_method.tex
\section{Method}
\subsection{Formulation}
\paragraph{Conditional Flow matching for audio-driven video generation}
Flow matching video generative models \cite{flowmatching, chen2025hunyuanvideo, wan} adopts a neural network to generate realistic video frames by modeling a timestep-dependent vector field that transports samples from a noise distribution to a target video distribution. Given a ground truth conditional video distribution $q(\boldsymbol{x} \vert \boldsymbol{c})$ where $\boldsymbol{x} \in \mathbb{R}^{t \times h \times w \times c}$ is the encoded video latent. $\boldsymbol{c} = \{y, a, \boldsymbol{x}_{\mathrm{ref}}\}, y \in \mathbb{R}^{m \times d_{\mathrm{text}}}, a \in \mathbb{R}^{n \times d_{\mathrm{audio}}}, \boldsymbol{x}_{\mathrm{ref}} \in \mathbb{R}^{t_{\mathrm{ref}} \times h\times w \times c}$ are the conditions, including the text prompt embedding, and the audio embedding, and the reference frames latent. Conditional flow matching defines a series of distributions by interpolating $q(\boldsymbol{x} \vert y, a)$ with a known trivial distribution (e.g. Gaussian noise) $p(\boldsymbol{x} \vert y, a)$ using a continuous variable $t \in [0. 1]$.

\begin{equation}
q_t(\boldsymbol{x} \vert y, a) = (1 - t) \cdot p(\boldsymbol{x} \vert y, a) + t \cdot q(\boldsymbol{x} \vert y, a).
\end{equation}

To be specific, a random variable $\boldsymbol{x}_t \sim q_t(\boldsymbol{x} \vert y, a)$ can be obtained by interpolating between $\boldsymbol{x}_0 \sim p(\boldsymbol{x} \vert y, a)$ and $\boldsymbol{x}_1 \sim q(\boldsymbol{x} \vert y, a)$ via $\boldsymbol{x}_t = (1 - t) \cdot \boldsymbol{x}_1 + t \cdot \boldsymbol{x}_0$. The generative model $\boldsymbol{v}_\theta(\cdot)$, parameterized by $\theta$, is trained to match a continuous velocity field $\boldsymbol{v}_\theta(\boldsymbol{x}_t \vert y, a) \simeq \frac{d \boldsymbol{x}_t}{d t}$. To achieve so, we adopt the conditional flow matching objective. 

\begin{equation}
    \mathcal{L}_{\mathrm{fm}} = \mathbb{E}_{t, \boldsymbol{x}_0, \boldsymbol{x}_1} \|\boldsymbol{v}_\theta(\boldsymbol{x}_t \vert y) - (\boldsymbol{x}_1 - \boldsymbol{x}_0)\|_2^2.
\end{equation}

 An ODE solver can be used to sample from a flow matching generative models.

\paragraph{Sparse-frame video dubbing}
Video dubbing localizes content by replacing original audio with translated speech while preserving visual authenticity. As formalized in this work, the task transforms a source video latent $\boldsymbol{x}_0 \in \mathbb{R}^{t \times h \times w \times c}$ and a target audio $a \in \mathbb{R}^{n \times d_{\mathrm{audio }}}$ into an output video where lip movements, facial expressions, and body dynamics synchronize organically with the new audio.
Traditional video dubbing techniques focus exclusively on oral region inpainting—editing lip movements while freezing head rotations, facial expressions, and body gestures \cite{li2024latentsync}. This creates immersion-breaking mismatches, as static body language contradicts emotional speech (e.g., a rigid posture during passionate dialogue). Sparse-frame video dubbing, illustrated in \cref{fig:teaser}, fundamentally redefines this process: it preserves only select keyframes $\boldsymbol{x}_{\mathrm{ref}}$ to anchor identity, emotional cadence, symbolic gestures, and camera trajectories—critical for visual continuity—while liberating full-body dynamics (facial expressions, head motions, body gestures) to organically synchronize with dubbed audio. As \cref{fig:teaser} demonstrates, this paradigm shift enables lifelike alignment where head turns follow speech rhythm and gestures amplify emotional tone—impossible with lip-only editing. Crucially, sparse-frame dubbing operates on infinite-length sequences, demanding generative continuation beyond short clips to maintain synchronization across extended durations, a capability unattainable with traditional frame-by-frame inpainting.

\subsection{Observation on naive solutions}
\input{fig/observation1}
\input{fig/observation}

This section investigates practical approaches for sparse-frame video dubbing using two baseline models: image-to-video (I2V) \cite{cui2024hallo3} and first-last-frame-to-video (FL2V) \cite{wan}. As illustrated in \cref{fig:observation1}, both methods exhibit critical limitations when generating long video sequences. The I2V approach operates by initializing the first chunk of the video from a single reference frame (e.g., the source video’s starting keyframe). For subsequent chunks, it uses only the last generated frame of the preceding chunk as the new reference. While this preserves motion flexibility, the lack of persistent anchoring to the original keyframes leads to accumulated errors: subtle discrepancies in identity (e.g., facial features gradually deviating from the source actor) and color tones (e.g., background hues shifting across chunks) compound over time, resulting in visible degradation. In contrast, the FL2V method conditions each chunk on both the start and end frames of the input segment, ensuring alignment with the source video’s reference poses. This eliminates accumulation errors but introduces a new problem: the model enforces rigid control by strictly replicating the reference frames at the corresponding timestamp. This contradicts the soft conditioning required for sparse-frame dubbing, where full body motions must dynamically adapt to audio cues.

Crucially, both methods suffer from abrupt inter-chunk transitions. Since they rely solely on static image conditions (e.g., a single frame for I2V or two fixed frames for FL2V), they lack momentum information that should carry over between chunks. These observations a highlight fundamental trade-off: I2V prioritizes motion fluidity at the expense of accumulated error, while FL2V prioritizes reference fidelity at the cost of motion naturalness.

\subsection{Audio-driven streaming video generator with reference frames}
\label{sec:pipeline}
\input{fig/pipeline}
\input{fig/mps}
To resolve the accumulated accumulated error in I2V models and the abrupt transitions in FL2V models, we build a audio-driven streaming human animation architecture. This framework employs context frames, defined as the trailing segment of each previously generated chunk, to propagate kinetic momentum into subsequent segments. By processing these frames through a diffusion transformer, the model sustains motion continuity. To eliminate accumulated errors, we adopt multiple reference frames dynamically sampled from the source video, similar to FL2V's multi-frame conditioning. These keyframes preserve critical visual attributes including identity, background details, camera trajectories, and stylistic elements. Crucially, unlike FL2V's rigid replication of reference frames at fixed positions, which suppresses natural motion. Our model has the potential to achieve soft conditioning as discussed in \cref{sec:strategies}.

The sketch for the model is shown in \cref{fig:pipeline}. The model consists an audio embedder \cite{wav2vec}, a video VAE, and diffusion transformer (DiT) \cite{dit}. We first introduce how our model is trained. In the followings, let the video latent refers to embedding derived by encoding pixel space video by using the video VAE. Similarly, let audio embedding denote the embedding computed by encoding a audio sequence using the audio embedder. To train the model, we do not need a dubbed video pair, but only a video with the audio track is needed. Given a source video. The reference frame is uniformly random sampled from the video. During training, the context frames are the first $4(t_c - 1) + 1$ frames of the source video. After VAE encoding, we derive reference frame latent $\boldsymbol{x}_{\mathrm{ref}} \in \mathbb{R}^{c \times 1 \times h \times w}$, the full source video latent $\boldsymbol{x}_{\mathrm{full}} \in \mathbb{R}^{c \times (t + t_c) \times h \times w}$, the context frames latent $\boldsymbol{x}_{\mathrm{context}} \in \mathbb{R}^{c \times t_c \times h \times w}$, and the subsequent frames latent $\boldsymbol{x}_0 \in \mathbb{R}^{c \times t \times h \times s}$ separately, where the full video latent is a combination of the context frames latent  $\boldsymbol{x}_{\mathrm{context}}$ and the subsequent frames latent $\boldsymbol{x}_0$, $\boldsymbol{x}_{\mathrm{full}} = \{\boldsymbol{x}_{\mathrm{context}}, \boldsymbol{x}_0\}$. The audio sequence in the source video is encoded to get embedding $a$. Without a loss of generality, we show the process to train at $t$ in conditional flow matching in this section. Unit gaussian noise is used as the trivial distribution. The noisy latent is derived by $\boldsymbol{x}_t = (1 - t) \cdot \boldsymbol{x}_1 + t \cdot \boldsymbol{x}_0, \boldsymbol{x}_1 \sim \mathcal{N}(\boldsymbol{0}, \mathbb{I})$ where $\boldsymbol{x}_1$ is gaussian noise that has the same dimensionality of $\boldsymbol{x}_0$. The DiT model formulates a field estimator $\boldsymbol{v}_\theta(\boldsymbol{x}_t \vert \boldsymbol{c})$ using condition $\boldsymbol{c} = \{y, a, \boldsymbol{x}_{\mathrm{ref}}, \boldsymbol{x}_{\mathrm{tran}}\}$. Specifically, we first concatenate the noisy latent $\boldsymbol{x}_t$ and the clean context frames $\boldsymbol{x}_{\mathrm{context}}$ in the temporal dimension to get $\boldsymbol{z}_1 \in \mathbb{R}^{c \times (t + t_c) \times h \times w}$. Then we pad the reference frame to the temporal length $t_c + t$ to get $\boldsymbol{z}_2 \in \mathbb{R}^{c \times (t + t_c) \times h \times w}$. Finally, we concatenate $\boldsymbol{z}_1$, $\boldsymbol{z}_2$ and a reference frame indicating mask $\boldsymbol{m} \in \mathbb{R}^{4 \times (t + t_c) \times h \times w}$ in channel dimension. Mathematically, the process is

\begin{equation}
\begin{aligned}
    \boldsymbol{z}_1 & = \mathrm{concat}((\boldsymbol{x}_{\mathrm{context}}, \boldsymbol{x}_t), 2) \\
    \boldsymbol{z}_2 & = \mathrm{concat}((\boldsymbol{x}_{\mathrm{ref}}, \mathbb{0}), 2) \\
    \boldsymbol{m}   & = \mathrm{concat}((\mathbb{1}, \mathbb{0}), 2) \\
    \boldsymbol{z}   & = \mathrm{concat}((\boldsymbol{z}_1, \boldsymbol{z}_2, \boldsymbol{m}), 1)
\end{aligned}
\end{equation}

where $\mathrm{concat}(\cdot)$ is the concatenation operator (e.g, $\mathrm{concat}((\boldsymbol{x}_{\mathrm{context}}, \boldsymbol{x}_t), 2)$ concatenates $\boldsymbol{x}_{\mathrm{context}}$ and $\boldsymbol{x}_t$ in the 2-nd dimension). $\mathbb{0}$ and $\mathbb{1}$ are zero and one tensors that have the dimensions $\mathbb{0} \in \mathbb{R}^{4 \times (t + t_c - 1) \times h \times w}, \mathbb{1} \in \mathbb{R}^{4 \times 1 \times h \times w}$. As depicted by \cref{fig:pipeline} \textbf{(right)}. Within our transformer model, there is an audio cross-attention module and an image cross-attention that achieves audio and reference image conditioning. The reference frame is processed by CLIP vision model \cite{openclip} to get the embedding $\boldsymbol{z}_{\mathrm{ref}}$ before feeding to the DiT. To train this model, we adopt the conditional flow matching objective \cite{flowmatching}.

\begin{equation}
    \mathcal{L}_{\mathrm{fm}} = \mathbb{E}_{t, \boldsymbol{x}_0, \boldsymbol{x}_1, \boldsymbol{c}} \|\boldsymbol{v}_\theta(\boldsymbol{x}_t \vert \boldsymbol{c}) - (\boldsymbol{x}_1 - \boldsymbol{x}_0)\|_2^2.
\end{equation}


Then, we briefly introduce the sampling method. An illustration is presented in \cref{fig:mps}. We generate the whole long video sequence by auto-regressively generating small video chunks. In the first video chunk, we use the first frame of the input video as the reference frame and no context frames are required. In the following video chunks, we use the last $4 (t_c - 1) + 1$ frames from the previous output video chunk as the context frames and use the first image of the input chunk as the reference frame.

\subsection{Soft conditioning}
\label{sec:strategies}
\paragraph{Control strength from the Reference frame}
This section explores strategies for achieving soft conditioning in sparse-frame video dubbing where the model must generate audio-aligned full-body motions without rigidly replicating reference frames that may conflict with dubbed speech. Meanwhile, we expect the model to have adaptive control strength: (1) when the reference is similar to the context frames, the control strength is weak such that the model produces diverse dynamics. (2) when the reference is very distinct to the context frames, the control strength is stronger to ensure better consistency in identity and background. Below, we show the analysis on the training strategies that match the requirements.

We initiate our investigation with Model M0, which samples reference frames uniformly at random from the current source input video chunk during training. As demonstrated in \cref{fig:observation2}, this approach exhibits excessive control strength. The model inappropriately duplicates reference content at arbitrary timestamps, such as replicating slapping on the forehead during emotionally neutral speech, disrupting audio-visual synchronization.

To systematically analyze how reference positioning during training governs control fidelity, we examine two granular dimensions: chunk-level positioning (selecting which temporal segment provides the reference) and frame-level positioning (choosing the specific frame within that segment). We train three model variants to isolate these effects. Model M1 samples references exclusively from the first or last frame of the input chunk. This strategy mirrors FL2V's rigidity, forcing the generated video to replicate reference poses precisely at chunk boundaries, even when they contradict the audio's emotional cadence. Model M2 samples references from temporally distant chunks (e.g., segments separated by >5 seconds). While this weakens control sufficiently to avoid replication, it introduces accumulated color and background errors over long sequences, indicating insufficient fidelity preservation. In contrast, Model M3 samples references from adjacent chunks (e.g., within 1 seconds of the input). This configuration achieves moderate control strength: references preserve identity and camera motion without exact duplication, while eliminating accumulated errors entirely.

Our experiments conclusively demonstrate that chunk-level distance is the dominant factor modulating control strength. Shorter temporal distances (as in M3) create an optimal equilibrium: they anchor visual consistency to the source while liberating facial expressions, head rotations, and body gestures to organically synchronize with audio. Longer distances (as in M2) degrade preservation capabilities, leading to destabilized outputs. Fixed boundary sampling (as in M1) prioritizes replication over expressiveness, stifling motion dynamics. Thus, M3’s near-chunk positioning emerges as the foundational strategy for soft conditioning—enabling faithful yet flexible video dubbing where motions breathe in harmony with speech. 
\paragraph{Camera control}
We explore ways to address the camera movement preservation in sparse-frame video dubbing in this section. The utilization of reference frames is providing a global control of camera trajectory. However, the detailed camera movement within each video chunk is not controlled and may contradict the source video. To resolve this, we use two plugin including SDEdit \cite{sdedit} and Uni3C \cite{cao2025uni3c}. SDEdit incorporates trajectory information by adding the source video to the initialize noise at a scale $t_0$. The denoising sampling process starts from $t= t_0$ instead of $t=1$. $\boldsymbol{x}_{t_0} = (1 - t_0) \cdot \boldsymbol{x}_1 + t_0 \cdot \boldsymbol{x}_0$
Uni3C injects camera movements by deploying a ControlNet-like architecture. A comparison between the methods is shown in our experiments.

%% file: fig/observation1.tex
\begin{figure*}[t]
  \centering
  \includegraphics[width=1.0\textwidth]{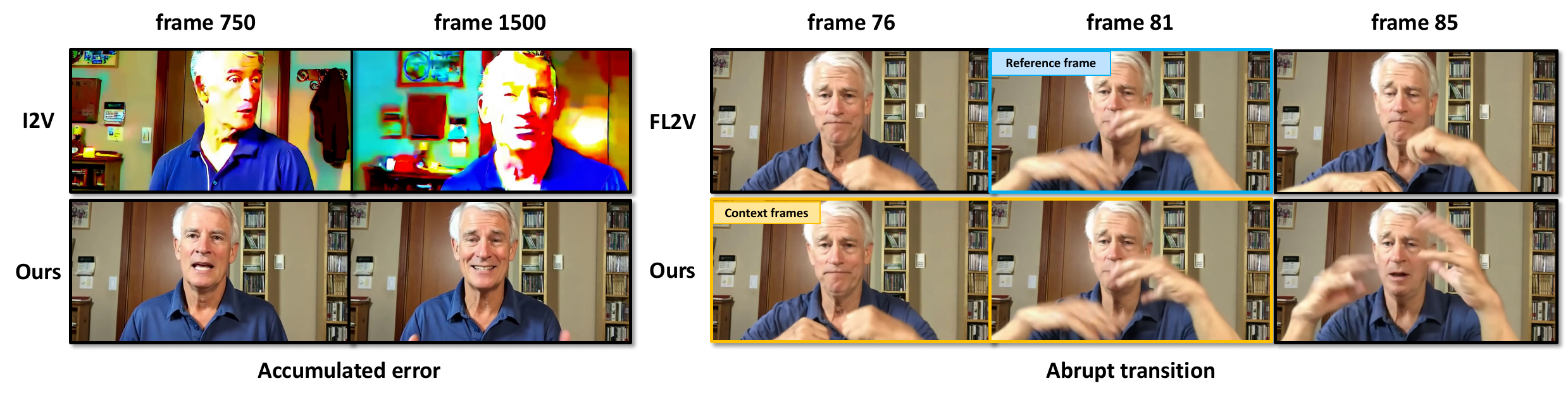}
  \caption{
    (\textbf{left}): I2V model accumulates error for long video sequences. (\textbf{right}): A new chunk starts from frame 82. FL2V model suffers from abrupt inter-chunk transitions. 
  }
  \label{fig:observation1}
\end{figure*}

%% file: fig/observation.tex
\begin{figure*}[t]
  \centering
  \includegraphics[width=1.0\textwidth]{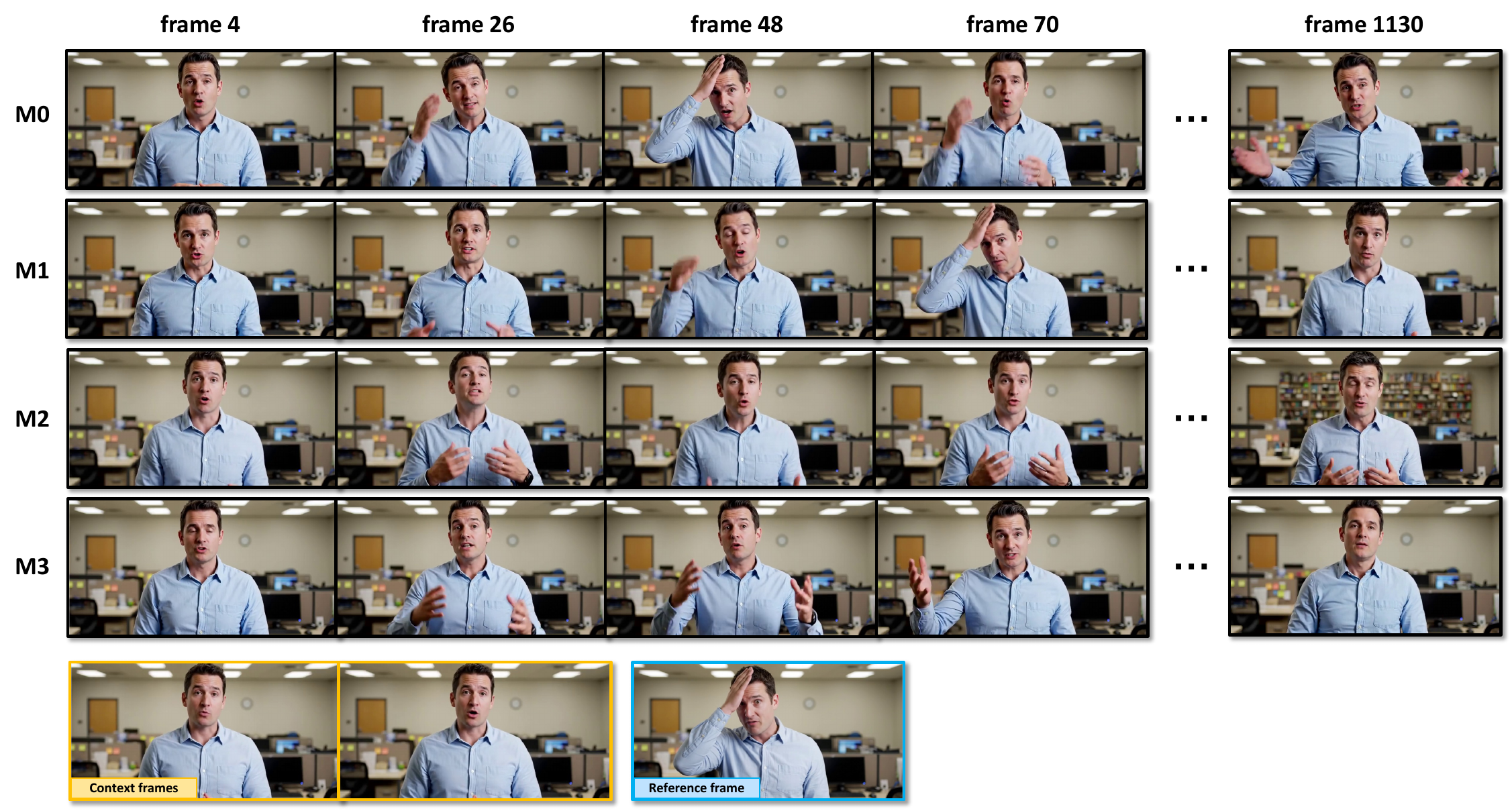}
  \caption{
    A visual comparison between the training reference positioning strategies. All video chunks are generated using the same context frames and the same reference frame shown in below. 
  }
  \label{fig:observation2}
\end{figure*}

%% file: fig/pipeline.tex
\begin{figure*}[t]
  \centering
  \includegraphics[width=1.0\textwidth]{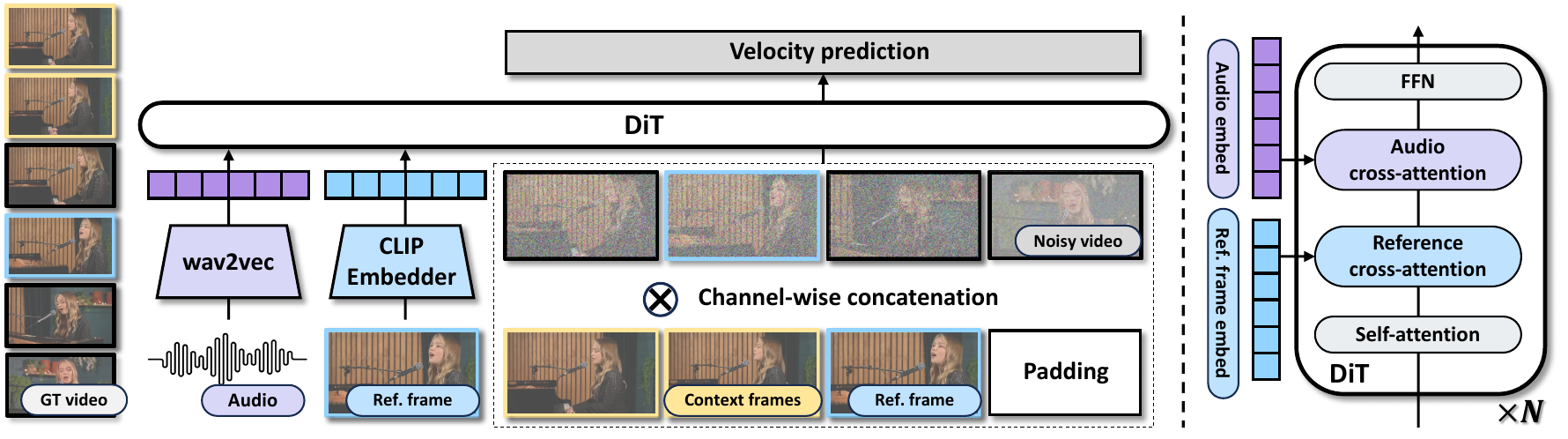}
  \caption{
    Visualization of InfiniteTalk pipeline. \textbf{Left}: The streaming model receives a audio, a reference frame, and context frames to denoise iteratively. \textbf{Right}: The architecture of the diffusion transformer. In addition to the traditional structures, each block includes an audio cross-attention layer and a reference cross-attention layer 
  }
  \label{fig:pipeline}
\end{figure*}

%% file: fig/mps.tex
\begin{figure*}[t]
  \centering
  \includegraphics[width=1.0\textwidth]{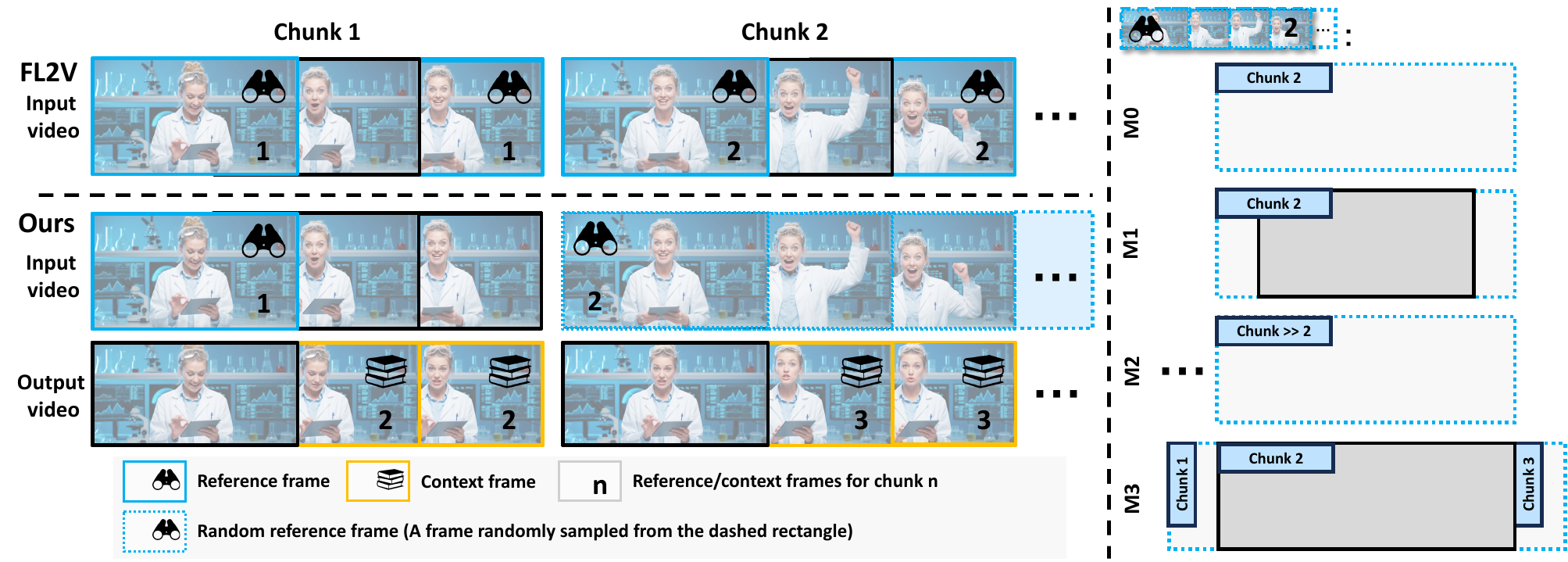}
  \caption{
  Visualization of reference frame conditioning strategies for video dubbing models. Top four rows: conditioning on input video frames. Bottom row: conditioning on generated video frames. \textbf{Left}: Image-to-video dubbing model with initial frame conditioning (I2V) and initial+terminal frame conditioning (IT2V). \textbf{Right}: Streaming dubbing model with four conditioning strategies. Within each category (left/right), all strategies share identical generated-video conditioning approaches.
  }
  \label{fig:mps}
\end{figure*}

%% file: sec/4_experiment.tex
\section{Experiment}
\paragraph{Implementation details}{
    We build our model based on MeiGen-MultiTalk \cite{kong2025let}, which includes a 14B parameters DiT that enables audio-driven image-to-video generation at multiple resolutions. We use wav2vec2 \cite{wav2vec2} as the audio embedder, and CLIP/H \cite{openclip} as the reference image embedder. Around 2,000 hours of video containing a talking person is collected as our training data. The model is trained with a 64 NVIDIA H100 80G cluster. The context frames includes 9 images, resulting in a $t_c = 3$ context frames latent. The video chunk length is 81. Our model will produce 72 frames each time when generating a long video auto-regressively.
}
\paragraph{Test datasets and evaluation metrics}{
    To rigorously evaluate our method across diverse scenarios, we utilize three benchmark datasets: HDTF \cite{hdtf} and CelebV-HQ \cite{celebvhq} (emphasizing facial dynamics) alongside EMTD \cite{emtd} (incorporating full-body movements). Following established dubbing evaluation protocols \cite{li2024latentsync, skyreelsaudio}, we construct a test set of 120 videos by randomly sampling 40 videos per dataset and permuting their audio channels (replacing original audio with mismatched tracks) to simulate real dubbing conditions. We use our model to perform long sequence video dubbing at $480\times 480$ resolution. The temporal length of generated results is the frame number of the the dubbing video input. 
    
    Performance is quantified through complementary automatic metrics and human evaluation. For objective assessment, we employ: Fréchet Inception Distance (FID) measuring per-frame visual quality; Fréchet Video Distance (FVD) evaluating inter-frame temporal coherence; SyncNet's Sync-C (confidence score) and Sync-D (lip distance) quantifying lip synchronization; and Cosine Similarity (CSIM) scoring identity preservation. To capture perceptual nuances beyond automated metrics, we conduct human studies where participants rate on 5 perspectives: gesture synchronization with audio prosody, head motion alignment to speech rhythm, lip synchronization precision, identity consistency, and overall perceptual naturalness. We receive 340 responses from 17 participants on all 40 video dubbing results from EMTD. 
}

\subsection{Quantitative experiments}
\input{fig/qualitative}

\input{tab/quantitative}
\input{tab/ablation}
\input{tab/human_evaluation}
We compare InfiniteTalk with both traditional video dubbing methods (including MuseTalk \cite{musetalk}, FantacyTalking \cite{wang2025fantasytalking}, and Hallo3 \cite{cui2024hallo3}) and audio-driven image-to-video models (including OmniAvatar \cite{gan2025omniavatar} and MultiTalk \cite{kong2025let}). To follow the pre-processing pipeline and accurately show the performance of the counterparts, we use their open-source weights and inference scripts in this experiment. 

The comparison with image-to-video models is shown in \cref{tab:quantitative-i2v}. InfiniteTalk outperforms the counterparts in lip synchronization by a large margin. Note that, there is a trade-off between synchronization metrics (Sync-C, Sync-D), visual quality matrics (FID, FVD) and identity preservation (CSIM). A trivial solution is that, if a method is copying the input as the output, it will achieve the best FID, FVD, and CSIM over all methods. When comparing InfiniteTalk with methods that have competitive synchronization performances, our method achieves very remarkable visual quality and identity preserving. As seen in \cref{tab:quantitative-dub} when comparing with traditional video dubbing methods. Since LatentSync \cite{li2024latentsync} and MuseTalk \cite{musetalk} is limited to edit the oral region, the rest of the video is kept unchanged, making their FID and FVD extremely good. Given that, the metrics are not showing the true visual quality difference. 

Currently, there is not automatic full body motion-audio alignment metric available. We find the music-motion alignment metric like beat consistency score fail to differentiate the performance of the methods. Meanwhile, Sync-C and Sync-D cannot accurately depict the lip synchronization when there is large head motion in the video.
For a comprehensive evaluation of full body motion alignment, we conduct a user study. 
The result is shown in \cref{tab:human_evaluation}. The participant is asked to rank the results (placing the best to the 1-st place, the worst to the 3-rd place) by the video dubbing methods including MuseTalk~\cite{musetalk}, LatentSync~\cite{li2024latentsync}, and our method. The number in the table shows the averaged ranking of the corresponding method. Benefited by the strong audio-driven human animation architecture and the sparse-frame video dubbing paradigm, our method achieves the best results in both lip and body motion synchronization. It demonstrates the limitation of traditional video dubbing methods that the editing region is restricted to mouth, resulting to misaligned body motion.

\subsection{Qualitative experiments}
\paragraph{Comparison with counterparts}
We conduct a visual comparison between our method and traditional video dubbing methods in \cref{fig:qualitative}. The first input example is a static video. It showcases when only editing mouth regions, traditional video dubbing methods cannot drive the head and body by the audio track. The following two inputs are dynamic videos. Compared to the counterparts, InfiniteTalk is not only able to generate plausible audio-aligned lip movements, but also synchronized face, head, and body movements with matched emotional expressions. As a new paradigm, sparse-frame video dubbing also demonstrates its necessity for modern audio-driven human animation video-to-video applications.

\input{fig/camera_control}
\paragraph{Camera control}
A visual comparison on the camera control methods are shown in \cref{fig:camera_control}. Using InfiniteTalk alone will not replicate the subtle camera movement of source video. With SDEdit~\cite{sdedit} or Uni3C~\cite{cao2025uni3c}, we can achieve fine-grained camera control. Comparing SDEdit with Uni3C. We find that with Uni3C, the model fails to preserve the video background. 


\subsection{Ablation study}
To systematically evaluate which training strategy performs the best in sparse-frame video dubbing, we conduct ablation studies comparing the four different reference frame positioning methods introduced in \cref{sec:strategies}. 
All ablated models are rigorously benchmarked using the 40-video test set sampled from EMTD under identical conditions.
As shown in \cref{tab:ablation}, our fine-grained reference frame positioning during training is the key to achieve reliable visual quality and audio-motion synchronization.

%% file: fig/qualitative.tex
\begin{figure*}[t]
  \centering
  \includegraphics[width=1.0\textwidth]{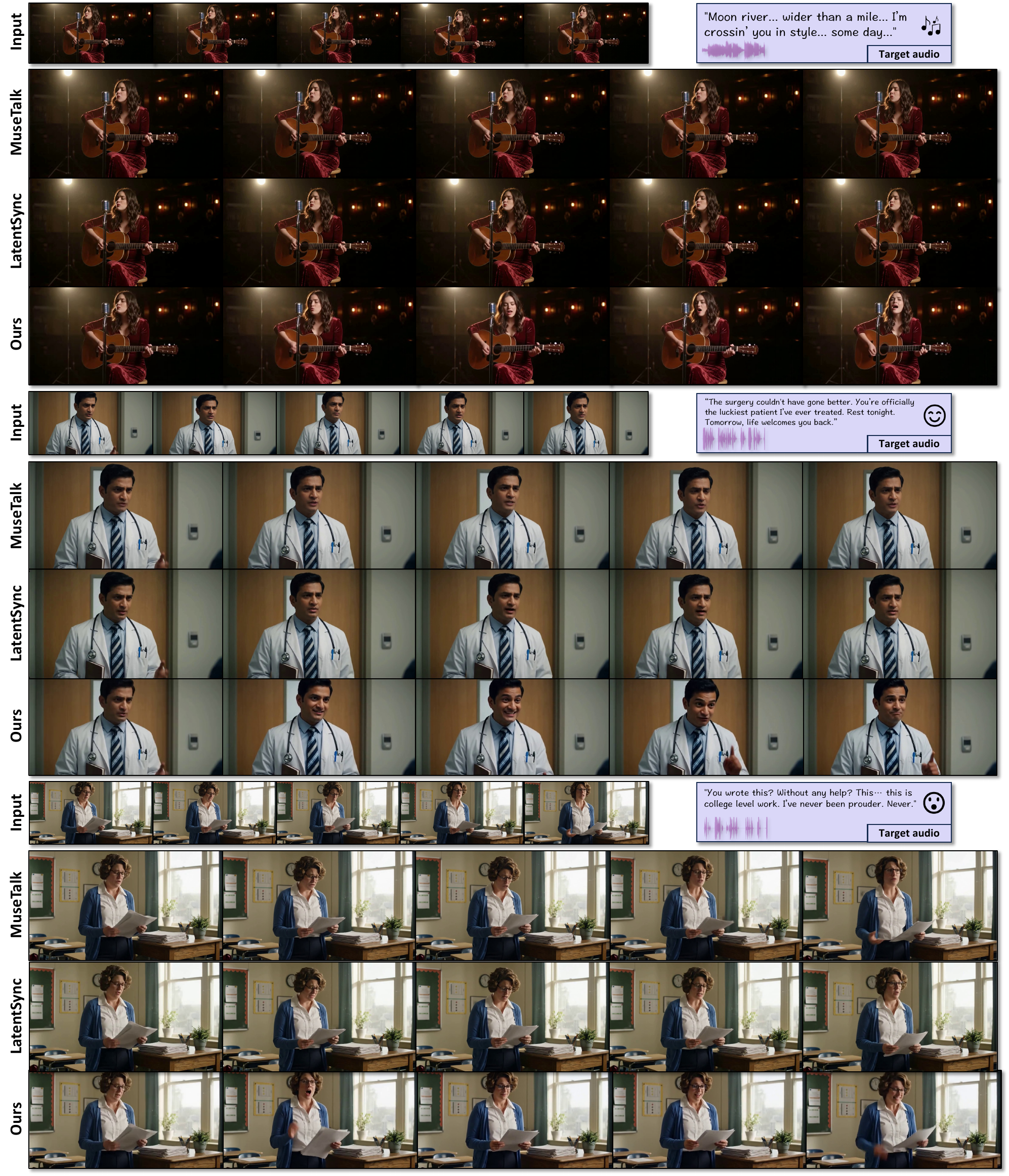}
  \caption{
    A visual comparison between the video dubbing methods. 
  }
  \label{fig:qualitative}
\end{figure*}

%% file: tab/quantitative.tex
\begin{table*}[t]
        \centering
        \begin{tabular}{c|c|ccccc}
        \toprule
            \multirow{2}{*}{Dataset} & \multirow{2}{*}{Model} & \multicolumn{4}{c}{Metrics} \\
            & & FID$\downarrow$ & FVD $\downarrow$ & Sync-C$\uparrow$ & Sync-D$\downarrow$ & CSIM$\uparrow$
            \\
            \midrule
            \multirow{3}{*}{HDTF} 
            & LatentSync        & 16.09 & \textbf{48.45}  & 8.99 & \textbf{6.36} & 0.916 \\
            & MuseTalk          & \textbf{14.20} & 49.13  & 7.17 & 7.90 & 0.933 \\
            & Ours              & 26.11 & 131.65 & \textbf{9.35} & 6.67 & 0.775 \\
            \midrule
            \multirow{3}{*}{CelebV-HQ} 
            & LatentSync        & 17.80 & \textbf{67.97}  & 6.90 & 7.33 & 0.869 \\
            & MuseTalk          & \textbf{17.62} & 72.07  & 4.16 & 9.86 & 0.857 \\
            & Ours              & 32.29 & 229.67 & \textbf{7.53} & \textbf{7.33} & 0.726 \\
            \midrule
            \multirow{3}{*}{EMTD} 
            & LatentSync        & \textbf{11.43} & 212.60 & 8.10 & \textbf{6.97} & 0.846 \\
            & MuseTalk          & 14.26 & \textbf{46.07}  & 5.35 & 9.28 & 0.825 \\
            & Ours              & 32.55 & 312.17 & \textbf{8.60} & 7.16 & 0.713\\
            \bottomrule
        \end{tabular}
        \caption{
        Quantitative comparisons our methods between the traditional video dubbing models.
        }
        \label{tab:quantitative-dub}
\end{table*}

\begin{table*}[t]
        \centering
        \begin{tabular}{c|c|ccccc}
        \toprule
            \multirow{2}{*}{Dataset} & \multirow{2}{*}{Model} & \multicolumn{4}{c}{Metrics} \\
            & & FID$\downarrow$ & FVD $\downarrow$ & Sync-C$\uparrow$ & Sync-D$\downarrow$ & CSIM $\uparrow$
            \\
            \midrule
            \multirow{5}{*}{HDTF} 
            & FantacyTalking    & 32.06 & \textbf{110.36} & 3.78 & 10.80 & 0.684 \\
            & Hallo3            & 36.48 & 144.65 & 7.20 & 8.61 & 0.674 \\
            & OmniAvatar        & 26.63 & 112.49 & 7.06 & 8.63 & 0.752 \\
            & MultiTalk         & 27.61 & 133.58 & 9.02 & 6.96 & 0.754 \\
            & Ours              & \textbf{27.14} & 132.54 & \textbf{9.18} & \textbf{6.84} & 0.751 \\
            \midrule
            \multirow{5}{*}{CelebV-HQ} 
            & FantacyTalking    & 37.53 & 237.58 & 2.93 & 10.79 & 0.654 \\
            & Hallo3            & 42.36 & 258.65 & 5.63 & 9.12 & 0.591 \\
            & OmniAvatar        & 37.41 & 250.67 & 5.88 & 8.68  & 0.703 \\
            & MultiTalk         & 34.79 & 230.41 & 7.25 & 7.70 & 0.711 \\
            & Ours              & \textbf{33.96} & \textbf{230.12} & \textbf{7.41} & \textbf{7.59} & 0.713 \\
            \midrule
            \multirow{5}{*}{EMTD} 
            & FantacyTalking    & 36.66 & \textbf{298.24} & 3.60 & 11.31 & 0.626 \\
            & Hallo3            & 44.71 & 326.94 & 5.68 & 9.56  & 0.512 \\
            & OmniAvatar        & \textbf{29.47} & 308.14 & 6.93 & 8.55 & 0.694 \\
            & MultiTalk         & 33.80 & 315.33 & 8.13 & 7.50 & 0.702 \\
            & Ours              & 33.27 & 314.68 & \textbf{8.34} & \textbf{7.36} & 0.709 \\
            \bottomrule
        \end{tabular}
        \caption{
        Quantitative comparisons between our method and audio-driven image-to-video models.
        }
        \label{tab:quantitative-i2v}
\end{table*}

%% file: tab/ablation.tex
\begin{table*}[t]
        \centering
        \begin{tabular}{c|cccc}
        \toprule
            Model & FID$\downarrow$ & FVD $\downarrow$ & Sync-C$\uparrow$ & Sync-D$\downarrow$
            \\
            \midrule
            Ours (M0)       & 32.69 & 322.04 & 8.51 & 7.31 \\
            Ours (M1)       & \textbf{32.21} & \textbf{307.21} & 7.96 & 8.11 \\
            Ours (M2)       & 42.17 & 376.53 & 8.23 & 7.44 \\
            Ours (M3)       & 32.55 & 312.17 & \textbf{8.60} & \textbf{7.16} \\
            \bottomrule
        \end{tabular}
        \caption{
        Ablation experiment results on EMTD.
        }
        \label{tab:ablation}
\end{table*} 

%% file: tab/human_evaluation.tex
\begin{table*}[t]
        \centering
        \begin{tabular}{c|cc}
        \toprule
            Model & Lip Sync.$\downarrow$ & Body Sync. $\downarrow$
            \\
            \midrule
            MuseTalk & 2.57 & - \\
            LatentSync & 2.32 & 1.92  \\
            Ours & \textbf{1.11} & \textbf{1.09} \\
            \bottomrule
        \end{tabular}
        \caption{
        Human evaluation between video dubbing methods on motion synchronization. We do not compare our method with MuseTalk in body synchronization because MuseTalk has exactly the same body movement to LatentSync.
        }
        \label{tab:human_evaluation}
\end{table*}

%% file: fig/camera_control.tex
\begin{figure*}[t]
  \centering
  \includegraphics[width=1.0\textwidth]{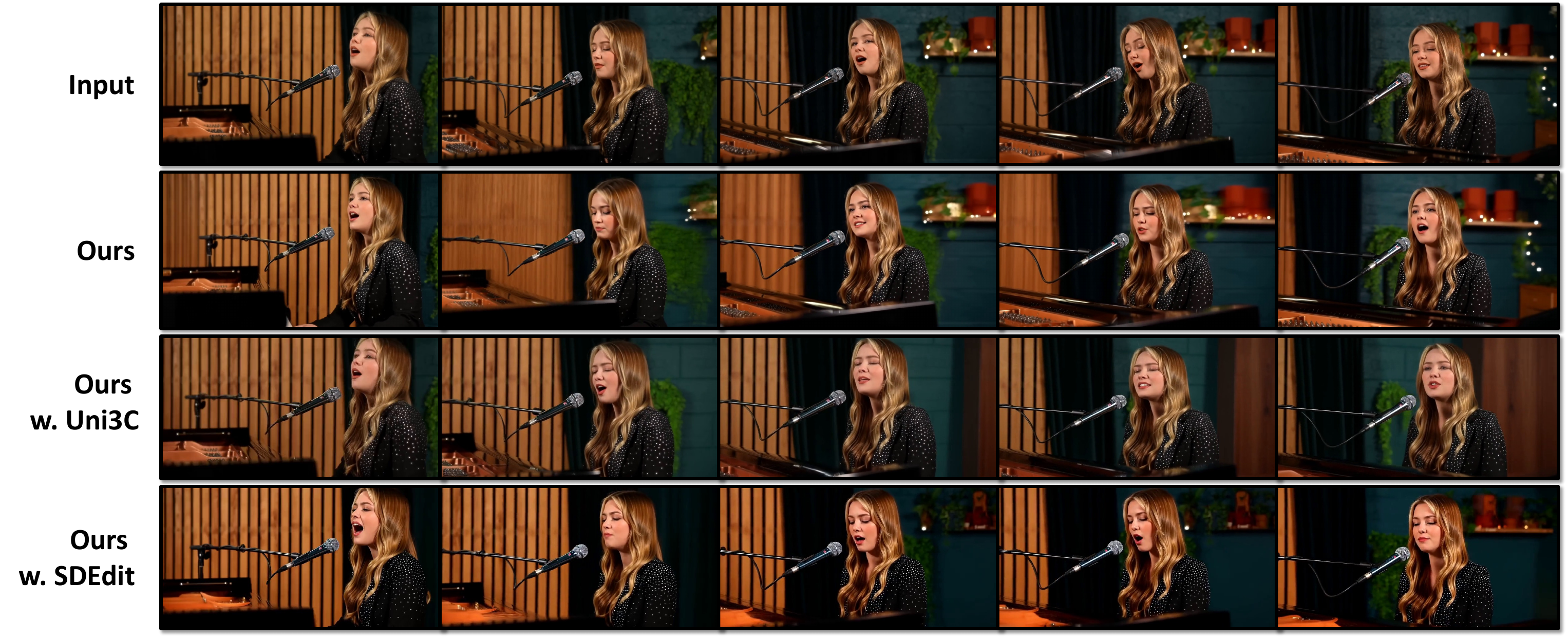}
  \caption{
    A visual comparison on the camera control.
  }
  \label{fig:camera_control}
\end{figure*}

%% file: sec/5_conclusion.tex
\section{Conclusion}

We introduce sparse-frame video dubbing, a novel paradigm for audio-driven video-to-video generation that employs reference keyframes to maintain emotional cadence and camera trajectories while liberating facial, head, and body dynamics to synchronize organically with dubbed audio. We propose InfiniteTalk, an audio-driven generator that overcomes critical limitations in long-form synthesis. By incorporating transient frame conditioning for seamless transitions, motion-provoking sampling to activate natural gestures, and adaptive camera control, InfiniteTalk achieves state-of-the-art lip, head, and body synchronization while eliminating identity drift and motion artifacts across extended sequences. Extensive validation confirms its superiority in producing natural, audio-aligned dynamics essential for immersive dubbed content.

%% file: main/iclr2025_conference.bbl
\begin{thebibliography}{54}
\providecommand{\natexlab}[1]{#1}
\providecommand{\url}[1]{\texttt{#1}}
\expandafter\ifx\csname urlstyle\endcsname\relax
  \providecommand{\doi}[1]{doi: #1}\else
  \providecommand{\doi}{doi: \begingroup \urlstyle{rm}\Url}\fi

\bibitem[Baevski et~al.(2020)Baevski, Zhou, Mohamed, and Auli]{wav2vec2}
Alexei Baevski, Henry Zhou, Abdelrahman Mohamed, and Michael Auli.
\newblock wav2vec 2.0: A framework for self-supervised learning of speech representations, 2020.
\newblock URL \url{https://arxiv.org/abs/2006.11477}.

\bibitem[Bigata et~al.(2025)Bigata, Mira, Bounareli, Stypułkowski, Vougioukas, Petridis, and Pantic]{keysync}
Antoni Bigata, Rodrigo Mira, Stella Bounareli, Michał Stypułkowski, Konstantinos Vougioukas, Stavros Petridis, and Maja Pantic.
\newblock Keysync: A robust approach for leakage-free lip synchronization in high resolution, 2025.
\newblock URL \url{https://arxiv.org/abs/2505.00497}.

\bibitem[Blattmann et~al.(2023{\natexlab{a}})Blattmann, Dockhorn, Kulal, Mendelevitch, Kilian, Lorenz, Levi, English, Voleti, Letts, Jampani, and Rombach]{stable-video-diffusion}
Andreas Blattmann, Tim Dockhorn, Sumith Kulal, Daniel Mendelevitch, Maciej Kilian, Dominik Lorenz, Yam Levi, Zion English, Vikram Voleti, Adam Letts, Varun Jampani, and Robin Rombach.
\newblock Stable video diffusion: Scaling latent video diffusion models to large datasets, 2023{\natexlab{a}}.

\bibitem[Blattmann et~al.(2023{\natexlab{b}})Blattmann, Rombach, Ling, Dockhorn, Kim, Fidler, and Kreis]{video-ldm}
Andreas Blattmann, Robin Rombach, Huan Ling, Tim Dockhorn, Seung~Wook Kim, Sanja Fidler, and Karsten Kreis.
\newblock Align your latents: High-resolution video synthesis with latent diffusion models.
\newblock In \emph{IEEE Conference on Computer Vision and Pattern Recognition ({CVPR})}, 2023{\natexlab{b}}.

\bibitem[Cao et~al.(2025)Cao, Zhou, Li, Liang, Yu, Wang, Xue, and Fu]{cao2025uni3c}
Chenjie Cao, Jingkai Zhou, shikai Li, Jingyun Liang, Chaohui Yu, Fan Wang, Xiangyang Xue, and Yanwei Fu.
\newblock Uni3c: Unifying precisely 3d-enhanced camera and human motion controls for video generation.
\newblock \emph{arXiv preprint arXiv:2504.14899}, 2025.

\bibitem[Chen et~al.(2023)Chen, Xia, He, Zhang, Cun, Yang, Xing, Liu, Chen, Wang, Weng, and Shan]{chen2023videocrafter1}
Haoxin Chen, Menghan Xia, Yingqing He, Yong Zhang, Xiaodong Cun, Shaoshu Yang, Jinbo Xing, Yaofang Liu, Qifeng Chen, Xintao Wang, Chao Weng, and Ying Shan.
\newblock Videocrafter1: Open diffusion models for high-quality video generation, 2023.

\bibitem[Chen et~al.(2024)Chen, Zhang, Cun, Xia, Wang, Weng, and Shan]{chen2024videocrafter2}
Haoxin Chen, Yong Zhang, Xiaodong Cun, Menghan Xia, Xintao Wang, Chao Weng, and Ying Shan.
\newblock Videocrafter2: Overcoming data limitations for high-quality video diffusion models, 2024.

\bibitem[Chen et~al.(2025{\natexlab{a}})Chen, Liang, Zhou, Huang, Ma, Tang, Lin, Zhou, and Lu]{chen2025hunyuanvideo}
Yi~Chen, Sen Liang, Zixiang Zhou, Ziyao Huang, Yifeng Ma, Junshu Tang, Qin Lin, Yuan Zhou, and Qinglin Lu.
\newblock Hunyuanvideo-avatar: High-fidelity audio-driven human animation for multiple characters.
\newblock \emph{arXiv preprint arXiv:2505.20156}, 2025{\natexlab{a}}.

\bibitem[Chen et~al.(2025{\natexlab{b}})Chen, Cao, Chen, Li, and Ma]{chen2025echomimic}
Zhiyuan Chen, Jiajiong Cao, Zhiquan Chen, Yuming Li, and Chenguang Ma.
\newblock Echomimic: Lifelike audio-driven portrait animations through editable landmark conditions.
\newblock In \emph{AAAI}, volume~39, pp.\  2403--2410, 2025{\natexlab{b}}.

\bibitem[Cherti et~al.(2023)Cherti, Beaumont, Wightman, Wortsman, Ilharco, Gordon, Schuhmann, Schmidt, and Jitsev]{openclip}
Mehdi Cherti, Romain Beaumont, Ross Wightman, Mitchell Wortsman, Gabriel Ilharco, Cade Gordon, Christoph Schuhmann, Ludwig Schmidt, and Jenia Jitsev.
\newblock Reproducible scaling laws for contrastive language-image learning.
\newblock In \emph{Proceedings of the IEEE/CVF Conference on Computer Vision and Pattern Recognition}, pp.\  2818--2829, 2023.

\bibitem[Cui et~al.(2024)Cui, Li, Zhan, Shang, Cheng, Ma, Mu, Zhou, Wang, and Zhu]{cui2024hallo3}
Jiahao Cui, Hui Li, Yun Zhan, Hanlin Shang, Kaihui Cheng, Yuqi Ma, Shan Mu, Hang Zhou, Jingdong Wang, and Siyu Zhu.
\newblock Hallo3: Highly dynamic and realistic portrait image animation with video diffusion transformer.
\newblock \emph{arXiv preprint arXiv:2412.00733}, 2024.

\bibitem[Esser et~al.(2021)Esser, Rombach, and Ommer]{vq-gan}
Patrick Esser, Robin Rombach, and Bjorn Ommer.
\newblock Taming transformers for high-resolution image synthesis.
\newblock In \emph{Proceedings of the IEEE/CVF conference on computer vision and pattern recognition}, pp.\  12873--12883, 2021.

\bibitem[Fei et~al.(2025)Fei, Jiang, Qiu, Gu, Zhang, Wang, Bai, Li, Fan, Chen, and Zhou]{skyreelsaudio}
Zhengcong Fei, Hao Jiang, Di~Qiu, Baoxuan Gu, Youqiang Zhang, Jiahua Wang, Jialin Bai, Debang Li, Mingyuan Fan, Guibin Chen, and Yahui Zhou.
\newblock Skyreels-audio: Omni audio-conditioned talking portraits in video diffusion transformers, 2025.
\newblock URL \url{https://arxiv.org/abs/2506.00830}.

\bibitem[Gan et~al.(2025)Gan, Yang, Zhu, Xue, and Hoi]{gan2025omniavatar}
Qijun Gan, Ruizi Yang, Jianke Zhu, Shaofei Xue, and Steven Hoi.
\newblock Omniavatar: Efficient audio-driven avatar video generation with adaptive body animation.
\newblock \emph{arXiv preprint arXiv:2506.18866}, 2025.

\bibitem[Ge et~al.(2023)Ge, Nah, Liu, Poon, Tao, Catanzaro, Jacobs, Huang, Liu, and Balaji]{pyoco}
Songwei Ge, Seungjun Nah, Guilin Liu, Tyler Poon, Andrew Tao, Bryan Catanzaro, David Jacobs, Jia-Bin Huang, Ming-Yu Liu, and Yogesh Balaji.
\newblock Preserve your own correlation: A noise prior for video diffusion models.
\newblock \emph{arXiv preprint arXiv:2305.10474}, 2023.

\bibitem[He et~al.(2022)He, Yang, Zhang, Shan, and Chen]{lvdm}
Yingqing He, Tianyu Yang, Yong Zhang, Ying Shan, and Qifeng Chen.
\newblock Latent video diffusion models for high-fidelity video generation with arbitrary lengths.
\newblock \emph{arXiv preprint arXiv:2211.13221}, 2022.

\bibitem[Ho et~al.(2022{\natexlab{a}})Ho, Chan, Saharia, Whang, Gao, Gritsenko, Kingma, Poole, Norouzi, Fleet, et~al.]{imagen-video}
Jonathan Ho, William Chan, Chitwan Saharia, Jay Whang, Ruiqi Gao, Alexey Gritsenko, Diederik~P Kingma, Ben Poole, Mohammad Norouzi, David~J Fleet, et~al.
\newblock Imagen video: High definition video generation with diffusion models.
\newblock \emph{arXiv preprint arXiv:2210.02303}, 2022{\natexlab{a}}.

\bibitem[Ho et~al.(2022{\natexlab{b}})Ho, Salimans, Gritsenko, Chan, Norouzi, and Fleet]{video-dm}
Jonathan Ho, Tim Salimans, Alexey Gritsenko, William Chan, Mohammad Norouzi, and David~J Fleet.
\newblock Video diffusion models.
\newblock \emph{arXiv preprint arXiv:2204.03458}, 2022{\natexlab{b}}.

\bibitem[Huang et~al.(2025)Huang, Wang, Zhao, Xu, Liu, and Chen]{huang2025bind}
Yubo Huang, Weiqiang Wang, Sirui Zhao, Tong Xu, Lin Liu, and Enhong Chen.
\newblock Bind-your-avatar: Multi-talking-character video generation with dynamic 3d-mask-based embedding router.
\newblock \emph{arXiv preprint arXiv:2506.19833}, 2025.

\bibitem[Ji et~al.(2024)Ji, Hu, Xu, Zhu, Lin, He, Zhang, Luo, Chen, Lin, et~al.]{ji2024sonic}
Xiaozhong Ji, Xiaobin Hu, Zhihong Xu, Junwei Zhu, Chuming Lin, Qingdong He, Jiangning Zhang, Donghao Luo, Yi~Chen, Qin Lin, et~al.
\newblock Sonic: Shifting focus to global audio perception in portrait animation.
\newblock \emph{arXiv preprint arXiv:2411.16331}, 2024.

\bibitem[Jiang et~al.(2024)Jiang, Liang, Yang, Lin, Zhong, and Zheng]{jiang2024loopy}
Jianwen Jiang, Chao Liang, Jiaqi Yang, Gaojie Lin, Tianyun Zhong, and Yanbo Zheng.
\newblock Loopy: Taming audio-driven portrait avatar with long-term motion dependency.
\newblock \emph{arXiv preprint arXiv:2409.02634}, 2024.

\bibitem[Kong et~al.(2025{\natexlab{a}})Kong, Gao, Zhang, Kang, Wei, Cai, Chen, and Luo]{kong2025let}
Zhe Kong, Feng Gao, Yong Zhang, Zhuoliang Kang, Xiaoming Wei, Xunliang Cai, Guanying Chen, and Wenhan Luo.
\newblock Let them talk: Audio-driven multi-person conversational video generation.
\newblock \emph{arXiv preprint arXiv:2505.22647}, 2025{\natexlab{a}}.

\bibitem[Kong et~al.(2025{\natexlab{b}})Kong, Li, Zhang, Gao, Yang, Wang, Zhang, Kang, Wei, Chen, et~al.]{kong2025dam}
Zhe Kong, Le~Li, Yong Zhang, Feng Gao, Shaoshu Yang, Tao Wang, Kaihao Zhang, Zhuoliang Kang, Xiaoming Wei, Guanying Chen, et~al.
\newblock Dam-vsr: Disentanglement of appearance and motion for video super-resolution.
\newblock \emph{arXiv preprint arXiv:2507.01012}, 2025{\natexlab{b}}.

\bibitem[Li et~al.(2024)Li, Zhang, Xu, Xie, Feng, Peng, and Xing]{li2024latentsync}
Chunyu Li, Chao Zhang, Weikai Xu, Jinghui Xie, Weiguo Feng, Bingyue Peng, and Weiwei Xing.
\newblock Latentsync: Audio conditioned latent diffusion models for lip sync.
\newblock \emph{arXiv preprint arXiv:2412.09262}, 2024.

\bibitem[Lin et~al.()Lin, Jiang, Liang, Zhong, Yang, Zheng, and Zheng]{lincyberhost}
Gaojie Lin, Jianwen Jiang, Chao Liang, Tianyun Zhong, Jiaqi Yang, Zerong Zheng, and Yanbo Zheng.
\newblock Cyberhost: A one-stage diffusion framework for audio-driven talking body generation.
\newblock In \emph{ICLR}.

\bibitem[Lin et~al.(2025)Lin, Jiang, Yang, Zheng, and Liang]{lin2025omnihuman}
Gaojie Lin, Jianwen Jiang, Jiaqi Yang, Zerong Zheng, and Chao Liang.
\newblock Omnihuman-1: Rethinking the scaling-up of one-stage conditioned human animation models.
\newblock \emph{arXiv preprint arXiv:2502.01061}, 2025.

\bibitem[Liu et~al.(2023)Liu, Gong, and Liu]{flowmatching}
Xingchao Liu, Chengyue Gong, and Qiang Liu.
\newblock Flow straight and fast: Learning to generate and transfer data with rectified flow, 2023.
\newblock URL \url{https://arxiv.org/abs/2209.03003}.

\bibitem[Meng et~al.(2022)Meng, He, Song, Song, Wu, Zhu, and Ermon]{sdedit}
Chenlin Meng, Yutong He, Yang Song, Jiaming Song, Jiajun Wu, Jun-Yan Zhu, and Stefano Ermon.
\newblock {SDE}dit: Guided image synthesis and editing with stochastic differential equations.
\newblock In \emph{International Conference on Learning Representations}, 2022.

\bibitem[Meng et~al.(2024)Meng, Zhang, Li, and Ma]{meng2024echomimicv2}
Rang Meng, Xingyu Zhang, Yuming Li, and Chenguang Ma.
\newblock Echomimicv2: Towards striking, simplified, and semi-body human animation.
\newblock \emph{arXiv preprint arXiv:2411.10061}, 2024.

\bibitem[Nichol \& Dhariwal(2021)Nichol and Dhariwal]{im-ddpm}
Alexander~Quinn Nichol and Prafulla Dhariwal.
\newblock Improved denoising diffusion probabilistic models.
\newblock In \emph{International Conference on Machine Learning}, pp.\  8162--8171. PMLR, 2021.

\bibitem[Peebles \& Xie(2022)Peebles and Xie]{dit}
William Peebles and Saining Xie.
\newblock Scalable diffusion models with transformers.
\newblock \emph{arXiv preprint arXiv:2212.09748}, 2022.

\bibitem[Rang~Meng(2025)]{emtd}
Yuming Li Chenguang~Ma Rang~Meng, Xingyu~Zhang.
\newblock Echomimicv2: Towards striking, simplified, and semi-body human animation, 2025.

\bibitem[Schneider et~al.(2019)Schneider, Baevski, Collobert, and Auli]{wav2vec}
Steffen Schneider, Alexei Baevski, Ronan Collobert, and Michael Auli.
\newblock wav2vec: Unsupervised pre-training for speech recognition, 2019.
\newblock URL \url{https://arxiv.org/abs/1904.05862}.

\bibitem[Singer et~al.(2022)Singer, Polyak, Hayes, Yin, An, Zhang, Hu, Yang, Ashual, Gafni, et~al.]{make-a-video}
Uriel Singer, Adam Polyak, Thomas Hayes, Xi~Yin, Jie An, Songyang Zhang, Qiyuan Hu, Harry Yang, Oron Ashual, Oran Gafni, et~al.
\newblock Make-a-video: Text-to-video generation without text-video data.
\newblock \emph{arXiv preprint arXiv:2209.14792}, 2022.

\bibitem[Song et~al.(2020)Song, Meng, and Ermon]{ddim}
Jiaming Song, Chenlin Meng, and Stefano Ermon.
\newblock Denoising diffusion implicit models.
\newblock \emph{arXiv preprint arXiv:2010.02502}, 2020.

\bibitem[Team(2024)]{genmo2024mochi}
Genmo Team.
\newblock Mochi 1.
\newblock \url{https://github.com/genmoai/models}, 2024.

\bibitem[Tian et~al.(2024{\natexlab{a}})Tian, Jiang, Yuan, Peng, and Wang]{var}
Keyu Tian, Yi~Jiang, Zehuan Yuan, Bingyue Peng, and Liwei Wang.
\newblock Visual autoregressive modeling: Scalable image generation via next-scale prediction.
\newblock 2024{\natexlab{a}}.

\bibitem[Tian et~al.(2024{\natexlab{b}})Tian, Wang, Zhang, and Bo]{tian2024emo}
Linrui Tian, Qi~Wang, Bang Zhang, and Liefeng Bo.
\newblock Emo: Emote portrait alive generating expressive portrait videos with audio2video diffusion model under weak conditions.
\newblock In \emph{ECCV}, pp.\  244--260. Springer, 2024{\natexlab{b}}.

\bibitem[Tian et~al.(2025)Tian, Hu, Wang, Zhang, and Bo]{tian2025emo2}
Linrui Tian, Siqi Hu, Qi~Wang, Bang Zhang, and Liefeng Bo.
\newblock Emo2: End-effector guided audio-driven avatar video generation.
\newblock \emph{arXiv preprint arXiv:2501.10687}, 2025.

\bibitem[Tu et~al.(2025)Tu, Pan, Huang, Han, Xing, Dai, Luo, Wu, and Yu-Gang]{stableavatar}
Shuyuan Tu, Yueming Pan, Yinming Huang, Xintong Han, Zhen Xing, Qi~Dai, Chong Luo, Zuxuan Wu, and Jiang Yu-Gang.
\newblock Stableavatar: Infinite-length audio-driven avatar video generation.
\newblock \emph{arXiv preprint arXiv:2508.08248}, 2025.

\bibitem[Wan et~al.(2025)Wan, Wang, Ai, Wen, Mao, Xie, Chen, Yu, Zhao, Yang, Zeng, Wang, Zhang, Zhou, Wang, Chen, Zhu, Zhao, Yan, Huang, Feng, Zhang, Li, Wu, Chu, Feng, Zhang, Sun, Fang, Wang, Gui, Weng, Shen, Lin, Wang, Wang, Zhou, Wang, Shen, Yu, Shi, Huang, Xu, Kou, Lv, Li, Liu, Wang, Zhang, Huang, Li, Wu, Liu, Pan, Zheng, Hong, Shi, Feng, Jiang, Han, Wu, and Liu]{wan}
Team Wan, Ang Wang, Baole Ai, Bin Wen, Chaojie Mao, Chen-Wei Xie, Di~Chen, Feiwu Yu, Haiming Zhao, Jianxiao Yang, Jianyuan Zeng, Jiayu Wang, Jingfeng Zhang, Jingren Zhou, Jinkai Wang, Jixuan Chen, Kai Zhu, Kang Zhao, Keyu Yan, Lianghua Huang, Mengyang Feng, Ningyi Zhang, Pandeng Li, Pingyu Wu, Ruihang Chu, Ruili Feng, Shiwei Zhang, Siyang Sun, Tao Fang, Tianxing Wang, Tianyi Gui, Tingyu Weng, Tong Shen, Wei Lin, Wei Wang, Wei Wang, Wenmeng Zhou, Wente Wang, Wenting Shen, Wenyuan Yu, Xianzhong Shi, Xiaoming Huang, Xin Xu, Yan Kou, Yangyu Lv, Yifei Li, Yijing Liu, Yiming Wang, Yingya Zhang, Yitong Huang, Yong Li, You Wu, Yu~Liu, Yulin Pan, Yun Zheng, Yuntao Hong, Yupeng Shi, Yutong Feng, Zeyinzi Jiang, Zhen Han, Zhi-Fan Wu, and Ziyu Liu.
\newblock Wan: Open and advanced large-scale video generative models.
\newblock \emph{arXiv preprint arXiv:2503.20314}, 2025.

\bibitem[Wang et~al.(2025)Wang, Wang, Jiang, Fan, Zhang, Qi, Zhao, and Xu]{wang2025fantasytalking}
Mengchao Wang, Qiang Wang, Fan Jiang, Yaqi Fan, Yunpeng Zhang, Yonggang Qi, Kun Zhao, and Mu~Xu.
\newblock Fantasytalking: Realistic talking portrait generation via coherent motion synthesis.
\newblock \emph{arXiv preprint arXiv:2504.04842}, 2025.

\bibitem[Wang et~al.(2023{\natexlab{a}})Wang, Chen, Ma, Zhou, Huang, Wang, Yang, He, Yu, Yang, Guo, Wu, Si, Jiang, Chen, Loy, Dai, Lin, Qiao, and Liu]{wang2023lavie}
Yaohui Wang, Xinyuan Chen, Xin Ma, Shangchen Zhou, Ziqi Huang, Yi~Wang, Ceyuan Yang, Yinan He, Jiashuo Yu, Peiqing Yang, Yuwei Guo, Tianxing Wu, Chenyang Si, Yuming Jiang, Cunjian Chen, Chen~Change Loy, Bo~Dai, Dahua Lin, Yu~Qiao, and Ziwei Liu.
\newblock Lavie: High-quality video generation with cascaded latent diffusion models, 2023{\natexlab{a}}.

\bibitem[Wang et~al.(2023{\natexlab{b}})Wang, Yu, and Zhang]{ddnm}
Yinhuai Wang, Jiwen Yu, and Jian Zhang.
\newblock Zero-shot image restoration using denoising diffusion null-space model.
\newblock \emph{The Eleventh International Conference on Learning Representations}, 2023{\natexlab{b}}.

\bibitem[Wei et~al.(2024)Wei, Yang, and Wang]{wei2024aniportrait}
Huawei Wei, Zejun Yang, and Zhisheng Wang.
\newblock Aniportrait: Audio-driven synthesis of photorealistic portrait animation.
\newblock \emph{arXiv preprint arXiv:2403.17694}, 2024.

\bibitem[Weijie~Kong \& Jie~Jiang(2024)Weijie~Kong and Jie~Jiang]{kong2024hunyuanvideo}
Zijian Zhang Rox Min Zuozhuo Dai Jin Zhou Jiangfeng Xiong Xin Li Bo Wu Jianwei Zhang Kathrina Wu Qin Lin Aladdin Wang Andong Wang Changlin Li Duojun Huang Fang Yang Hao Tan Hongmei Wang Jacob Song Jiawang Bai Jianbing Wu Jinbao Xue Joey Wang Junkun Yuan Kai Wang Mengyang Liu Pengyu Li Shuai Li Weiyan Wang Wenqing Yu Xinchi Deng Yang Li Yanxin Long Yi Chen Yutao Cui Yuanbo Peng Zhentao Yu Zhiyu He Zhiyong Xu Zixiang Zhou Zunnan Xu Yangyu Tao Qinglin Lu Songtao Liu Daquan Zhou Hongfa Wang Yong Yang Di Wang Yuhong~Liu Weijie~Kong, Qi~Tian and along with Caesar~Zhong Jie~Jiang.
\newblock Hunyuanvideo: A systematic framework for large video generative models, 2024.
\newblock URL \url{https://arxiv.org/abs/2412.03603}.

\bibitem[Xing et~al.(2023)Xing, Xia, Liu, Zhang, Zhang, He, Liu, Chen, Cun, Wang, et~al.]{make-your-vid}
Jinbo Xing, Menghan Xia, Yuxin Liu, Yuechen Zhang, Yong Zhang, Yingqing He, Hanyuan Liu, Haoxin Chen, Xiaodong Cun, Xintao Wang, et~al.
\newblock Make-your-video: Customized video generation using textual and structural guidance.
\newblock \emph{arXiv preprint arXiv:2306.00943}, 2023.

\bibitem[Xu et~al.(2024)Xu, Li, Su, Shang, Zhang, Liu, Wang, Yao, and Zhu]{xu2024hallo}
Mingwang Xu, Hui Li, Qingkun Su, Hanlin Shang, Liwei Zhang, Ce~Liu, Jingdong Wang, Yao Yao, and Siyu Zhu.
\newblock Hallo: Hierarchical audio-driven visual synthesis for portrait image animation.
\newblock \emph{arXiv preprint arXiv:2406.08801}, 2024.

\bibitem[Yan et~al.(2021)Yan, Zhang, Abbeel, and Srinivas]{VideoGPT}
Wilson Yan, Yunzhi Zhang, Pieter Abbeel, and Aravind Srinivas.
\newblock Videogpt: Video generation using vq-vae and transformers.
\newblock \emph{arXiv preprint arXiv:2104.10157}, 2021.

\bibitem[Yang et~al.(2024)Yang, Teng, Zheng, Ding, Huang, Xu, Yang, Hong, Zhang, Feng, et~al.]{yang2024cogvideox}
Zhuoyi Yang, Jiayan Teng, Wendi Zheng, Ming Ding, Shiyu Huang, Jiazheng Xu, Yuanming Yang, Wenyi Hong, Xiaohan Zhang, Guanyu Feng, et~al.
\newblock Cogvideox: Text-to-video diffusion models with an expert transformer.
\newblock \emph{arXiv preprint arXiv:2408.06072}, 2024.

\bibitem[Zhang et~al.(2025)Zhang, Zhong, Liu, Chen, Wu, Zeng, Zhan, He, Huang, and Zhou]{musetalk}
Yue Zhang, Zhizhou Zhong, Minhao Liu, Zhaokang Chen, Bin Wu, Yubin Zeng, Chao Zhan, Yingjie He, Junxin Huang, and Wenjiang Zhou.
\newblock Musetalk: Real-time high-fidelity video dubbing via spatio-temporal sampling, 2025.
\newblock URL \url{https://arxiv.org/abs/2410.10122}.

\bibitem[Zhang et~al.(2021)Zhang, Li, Ding, and Fan]{hdtf}
Zhimeng Zhang, Lincheng Li, Yu~Ding, and Changjie Fan.
\newblock Flow-guided one-shot talking face generation with a high-resolution audio-visual dataset.
\newblock In \emph{Proceedings of the IEEE/CVF Conference on Computer Vision and Pattern Recognition}, pp.\  3661--3670, 2021.

\bibitem[Zhou et~al.(2022)Zhou, Wang, Yan, Lv, Zhu, and Feng]{magicvideo}
Daquan Zhou, Weimin Wang, Hanshu Yan, Weiwei Lv, Yizhe Zhu, and Jiashi Feng.
\newblock Magicvideo: Efficient video generation with latent diffusion models.
\newblock \emph{arXiv preprint arXiv:2211.11018}, 2022.

\bibitem[Zhu et~al.(2022)Zhu, Wu, Zhu, Jiang, Tang, Zhang, Liu, and Loy]{celebvhq}
Hao Zhu, Wayne Wu, Wentao Zhu, Liming Jiang, Siwei Tang, Li~Zhang, Ziwei Liu, and Chen~Change Loy.
\newblock {CelebV-HQ}: A large-scale video facial attributes dataset.
\newblock In \emph{ECCV}, 2022.

\end{thebibliography}
